\documentclass[10pt,letterpaper]{article}
\usepackage{authblk}
\usepackage{ccn}
\usepackage{pslatex}
\usepackage{apacite}
\usepackage{hyperref}
\usepackage{natbib}
\usepackage{lineno}
\usepackage{nicefrac}
\usepackage{booktabs} 
\usepackage{tabularx}
\usepackage{graphicx}
\usepackage[table]{xcolor} 
\usepackage{algorithm}
\usepackage{algorithmic}
\usepackage{cancel}
\usepackage{makecell}
\usepackage{multirow}

\usepackage{amsmath,amsfonts,bm}

\def\eqref#1{equation~\ref{#1}}

\def\1{\bm{1}}

\def\vh{{\bm{h}}}

\def\vr{{\bm{r}}}

\def\vv{{\bm{v}}}

\def\vx{{\bm{x}}}

\def\vz{{\bm{z}}}

\def\mW{{\bm{W}}}

\DeclareMathAlphabet{\mathsfit}{\encodingdefault}{\sfdefault}{m}{sl}
\SetMathAlphabet{\mathsfit}{bold}{\encodingdefault}{\sfdefault}{bx}{n}

\newcommand{\E}{\mathbb{E}}

\DeclareMathOperator{\Tr}{Tr}

\title{Langevin Flows for Modeling Neural Latent Dynamics}
 
\author[1]{\large \bf Yue Song}
\author[2]{\large \bf T. Anderson Keller} 
\author[1]{\large \bf Yisong Yue} 
\author[1]{\large \bf Pietro Perona} 
\author[3]{\large \bf Max Welling} 

\affil[1]{ Caltech, Vision Lab, Pasadena, CA} 
\affil[2]{ The Kempner Institute for the Study of Natural and Artificial Intelligence, Harvard University\\
 Cambridge, MA}
\affil[3]{ University of Amsterdam, Institute for Informatics, Amsterdam, The Netherlands}

\begin{document}

\maketitle

\section{Abstract}
{
\bf Neural populations exhibit latent dynamical structures that drive time-evolving spiking activities, motivating the search for models that capture both intrinsic network dynamics and external unobserved influences. In this work, we introduce LangevinFlow, a sequential Variational Auto-Encoder where the time evolution of latent variables is governed by the underdamped Langevin equation.  Our approach incorporates physical priors — such as inertia, damping, a learned potential function, and stochastic forces — to represent both autonomous and non-autonomous processes in neural systems. Crucially, the potential function is parameterized as a network of locally coupled oscillators, biasing the model toward oscillatory and flow-like behaviors observed in biological neural populations. Our model features a recurrent encoder, a one-layer Transformer decoder, and Langevin dynamics in the latent space. Empirically, our method outperforms state-of-the-art baselines on synthetic neural populations generated by a Lorenz attractor, closely matching ground-truth firing rates. On the Neural Latents Benchmark (NLB), the model achieves superior held-out neuron likelihoods (bits per spike) and forward prediction accuracy across four challenging datasets. It also matches or surpasses alternative methods in decoding behavioral metrics such as hand velocity. Overall, this work introduces a flexible, physics-inspired, high-performing framework for modeling complex neural population dynamics and their unobserved influences. Code is available at \url{https://github.com/KingJamesSong/LangevinFlow_CCN}.



}
\begin{quote}
\small
\textbf{Keywords:} 
neural population dynamics; variational auto-encoders; latent variable models
\end{quote}

\section{Introduction}

Neural populations have been demonstrated to possess an underlying dynamical structure which drives the time evolution of population spiking activities~\citep{shenoy2013cortical,vyas2020computation}. Uncovering these underlying latent `factors' governing neural variability has become a goal of increasing interest in the neuroscience community. Such factors have been shown to be predictive of held-out neurons, future neural dynamics, and even behavior~\citep{GALLEGO2017978}. Recent works in this field have emphasized the importance of being able to model both internal deterministic dynamics, and potentially unobserved external influences (such as input
from sensory areas, or stochastic influences from other unmeasured brain regions). In established frameworks such as AutoLFADS~\citep{autolfads}, such influences have been captured by separately inferred control variables which modulate the dynamics of the inferred latent variables. Separate work has further modeled neural activity and particularly decision-making, through the use of learned potential functions that shape attractor-like population dynamics~\citep{engel}. Their work revealed a single decision variable embedded in a higher-dimensional population code, where heterogeneous neuronal firing could be explained by diverse tuning to the same latent process. Notably, the notion of an attractor mechanism aligns with the concept of a potential landscape, wherein neural trajectories evolve within an energy basin that facilitates stable or quasi-stable states. In parallel, recent developments in Transformer architectures~\citep{ye2021representation,ye2024neural} offer a promising avenue for neural data modeling by capturing long-range dependencies and global context across entire sequences -- complementing traditional methods that focus on local temporal interactions.

Drawing from physics, the Langevin equation is a stochastic differential equation which describes a system driven by both deterministic forces and stochastic environmental influences. We propose that the Langevin equation naturally integrates the key ingredients highlighted in prior studies: intrinsic (autonomous) dynamics, unobserved external or stochastic influences, and a potential function to shape attractor-like behavior. Specifically, we introduce a novel latent variable model for neural data that leverages underdamped Langevin dynamics to describe the time evolution of latent factors. This model includes terms representing inertia, damping, a potential function, and stochastic forces arising from both internal and external sources. Crucially, the potential function in our model is parameterized as a network of locally coupled oscillators, inducing a bias towards oscillatory and flow-like dynamics previously observed in neural latent activity~\citep{rotation}. This formulation captures the autonomous dynamics inherent to neural systems, providing a principled way to model both the stability and variability observed in neural responses. The oscillatory potential function also mirrors the emergence of cortical rhythms and traveling waves that have been linked to critical computational roles such as information integration, synchronization, and flexible sensorimotor processing~\citep{ermentrout2001traveling, buzsaki2006rhythms}. 

We train the model as a sequential Variational Auto-Encoder (VAE)~\citep{vae} with a recurrent encoder and a small one-layer transformer serving as the generative map from latent variables to neural spike rates. The recurrent encoder effectively captures local temporal dependencies in the neural data, while the Transformer decoder is employed to harness global context. By attending to the entire latent sequence, the Transformer refines firing rate predictions through integrating information from all timesteps, ensuring that long-range interactions and subtle dynamical patterns are well captured. This combination allows the model to capture complex temporal patterns and spatial correlations within the neural population data. Empirically, we first show the efficacy of our LangevinFlow on synthetic neural population data generated from a Lorenz attractor system, where our method is able to predict the firing rates closer to the ground truth than existing competitive baselines. We then demonstrate state-of-the-art performance on the Neural Latents Benchmark (NLB)~\citep{pei2021neural}, achieving superior results in modeling held-out neuron likelihoods (co-smoothing, bits per spike) and forward prediction accuracy across all four benchmark datasets (\texttt{MC\_Maze}, \texttt{MC\_RTT}, \texttt{Area2\_Bump}, and \texttt{DMFC\_RSG}), sampled at both $5$ and $20$ ms. The model also performs comparably or better in decoding behavioral metrics such as hand velocity. Notably, the time evolution of latent representations reveals smooth spatiotemporal wave dynamics, which is reminiscent of traveling waves observed in cortical activity~\citep{muller2018cortical}. This suggests that our coupled oscillator potential might capture key computational principles underlying neural information integration. Ultimately, we present this Langevin dynamics framework for neural data modeling, which incorporates inductive biases from physical principles and accounts for unobserved influences through its inherent stochastic dynamics. This general framework also allows for the flexible design of potential functions, opening up new doors for experimentation with latent dynamical systems.

\section{Related Work}

Neural population modeling has emerged as a key area in computational neuroscience, primarily driven by technological advances that now allow us to simultaneously record from hundreds or even thousands of neurons~\citep{stevenson2011advances}. Rather than focusing on individual neurons in isolation, population-level analyses seek to uncover the collective dynamics that shape brain function. These methods aim to capture moment-to-moment variability~\citep{churchland2006neural,ecker2010decorrelated}, shed light on network-wide interactions~\citep{cohen2011measuring,saxena2019towards}, and relate neural activity to behavior in real time~\citep{gallego2018cortical,gallego2020long,dabagia2023aligning} — all of which are central goals for both fundamental neuroscience research and applied domains such as brain-computer interfaces~\citep{sussillo2016making,karpowicz2022stabilizing}.

Early approaches to analyzing population neural recordings primarily focused on relatively simple statistical or latent-variable methods. Among the most widely used are linear and switching linear dynamical systems (LDS and SLDS)~\citep{macke2011empirical,kao2015single,gao2016linear,linderman2017bayesian}, which model neural population activity via linear state transitions (or piecewise linear segments) and emissions. Gaussian process-based approaches~\citep{yu2008gaussian,zhao2017variational,wu2017gaussian,duncker2018temporal} impose smoothness assumptions on latent factors and allow flexible, nonparametric modeling. However, the need for trial-averaging and the limited expressiveness of linear or Gaussian process latent variables can miss richer structures inherent in neural data, particularly during dynamic and nonlinear brain computations.
To overcome these limitations, recurrent neural network (RNN)-based methods have emerged as powerful tools to capture the non-linear dynamics~\citep{zhao2016interpretable,duncker2019learning}. One seminal work in this space is Latent Factor Analysis via Dynamical Systems (LFADS)~\citep{lfads}, which utilizes RNNs to model autonomous dynamics in single trials of spiking activity. LFADS infers latent trajectories that explain observed neural variability and has demonstrated impressive gains over traditional baselines. Subsequent work such as AutoLFADS~\citep{autolfads} refined this framework by allowing the model to separately infer putative “control” inputs, thereby accounting for unobserved external influences (\emph{e.g.,} sensory input or cognitive factors) that modulate neural dynamics. Following the advances in machine learning, recent work has begun exploring Transformer-based architectures for neural data. Transformers process input tokens in parallel, enabling potentially faster training and inference compared to sequential RNNs. Their success in large-scale language tasks has motivated adaptations such as the Neural Data Transformer (NDT)~\citep{ye2021representation} which modifies the Transformer encoder for neural spiking data, the improved version NDT2~\citep{ye2024neural} which further improves scaling across heterogeneous
contexts, and POYO~\citep{azabou2024unified} which leverages both cross-attention and PerceiverIO~\citep{jaegle2022perceiver} to construct a latent tokenization method for neural population activities. Other recent methods include~\citet{kudryashova2025band,pals2024inferring}.

The most relevant methods to our work are AutoLFADS~\citep{autolfads} and NDT~\citep{ye2021representation}. AutoLFADS and LFADS employ RNNs as the encoder and decoder networks, and the temporal dynamics are given by the hidden states, while NDT uses Transformers to encode the spiking data and additionally adopts masked modeling methodology to learn the context information. By contrast, our LangevinFlow employs a recurrent encoder, an oscillatory potential to enforce Langevin dynamics to the time evolution of latent variables, and a single Transformer layer to decode the entire variable sequence to firing rates.



\section{Methodology}

In this section, we first introduce the underdamped Langevin equation, then present the sequential VAE framework, followed by the derivation and analysis of how Langevin dynamics evolve in the posterior flow of latent variables. Finally, we discuss the model architecture and the training algorithm.

\subsection{Underdamped Langevin Equation}

We seek to build a latent variable model which integrates the desired beneficial inductive biases (intrinsic dynamics, stochastic influences, and an attractor-like potential function) in a principled manner. From the physics literature, a canonical abstract model of a system interacting with its environment is the Langevin equation:
\begin{equation}
\begin{aligned}
    \frac{\partial \vz}{\partial t} = \vv,\ \ 
     m\frac{\partial \vv}{\partial t}  = F(\vz) - m\gamma \vv + \sqrt{2m\gamma k_B \tau} \boldsymbol{\eta}(t)
    \label{eq:ud_langevin}
\end{aligned} 
\end{equation}
where $\vz(t)$ denotes the ($d$-dimensional) state of the system at time $t$, $\vv$ represents the associated velocity, $m$ is a diagonal matrix of masses, $F$ is the set of internal forces acting on the system (as a function of its state), $\gamma$ is the damping (or friction) coefficient, $k_B$ is the Boltzmann constant, $\tau$ is the temperature, and $\boldsymbol{\eta}(t)$ represents high-dimensional Gaussian white noise modeling the thermal fluctuation.

One method for defining the force field $F$ is in terms of the gradient of a scalar potential function $F(\mathbf{z}) = -\nabla_\mathbf{z} U(\mathbf{z})$. This formulation allows for the description of many well-known physical systems which have intrinsic dynamics. One abstraction of neural dynamics is that of a network of locally coupled oscillators~\citep{Diamant1969, ermentrout1984frequency}, which admits a particularly simple potential function:
\begin{equation}
    U(\vz) = \vz^T \frac{\mW_\vz}{||\mW_\vz||_2} \vz    
\end{equation} 
where $\mW_\vz \in \mathbb{R}^{d \times d}$ is the symmetric matrix of coupling coefficients between the individual oscillators. For a locally coupled system, this matrix reduces to a convolution operator in the Toeplitz form. Driven by this coupled oscillator potential, the time evolution of the latent state vector $\vz$ will have smooth spatiotemporal oscillatory dynamics (see Fig.~\ref{fig:waves}). 

\subsection{Sequential Variational Auto-Encoder}
To leverage the Langevin equation in a latent variable model of neural data, we assert that the sequence of observed spikes $\bar{\vx}$ is Poisson distributed according to the firing rate $\bar{\vr}$:
\begin{equation}
    p(\bar{\vx} | \bar{\vr}) = \sum_{t=0}^T \text{Poisson}(\vx_t | \vr_t)
\end{equation}
The firing rate is predicted by a decoder which takes as input the latent state variables, detailed later. For the input sequence $\bar{\vx}$, latent samples $\bar{\vz}$, and sample velocities $\bar{\vv}$, we further assert the following factorization of their joint distribution:
\begin{equation}
\resizebox{.90\linewidth}{!}{$
\begin{aligned}
    &p(\bar{\vx}, \bar{\vz}, \bar{\vv}) =p(\vv_0) p(\vz_0)  \prod_{t=1}^T p(\vv_t)p(\vz_t) \prod_{t=0}^T p(\vx_t|\vz_t,\vv_t) \\
     &=p(\vv_0) p(\vz_0)  \prod_{t=1}^T p(\vv_t)\delta(\vz_t - f_\vz(\vz_{t-1},\vv_{t-1})) \prod_{t=0}^T p(\vx_t|\vz_t,\vv_t) 
\end{aligned}
$}
\end{equation}
where $\delta(\cdot)$ denotes the Dirac $\delta$-function, and $f_\vz$ denotes the coupled Hamiltonian update which is introduced later in Eq.~(\ref{eq:hamilton_update}). Since $\vz_t$ and $\vv_t$ are coupled, the update to $\vz$ is deterministic. We thus only define $\vz_0$ and use $\delta$-functions to represent the later deterministic transformations. Here $p(\vz_0)$ and $p(\vv_t)$ are both standard Normal distributions, and $p(\vx_t|\vz_t,\vv_t)$ defines the mapping from latents to observations. 

We employ the framework of Variational Autoencoders (VAEs)~\citep{vae}, extended to sequential data, to perform inference over latent variables in this generative model. The goal of learning is to optimize the parameters of the following set of approximate posterior distributions:
\begin{equation}
\resizebox{.90\linewidth}{!}{$
\begin{aligned}
    &q_\theta (\bar{\vz},\bar{\vv} | \bar{\vx}) = q_\theta (\vz_0,\vv_0|\vx_0) q(\vz_{1:T}, \vv_{1:T}|\vz_0, \vv_0)\\
    &=q(\vz_0|\vx_0) q(\vv_0|\vx_0)  \prod_{t=1}^{T} q(\vz_{t}, \vv_t|\vz_{t-1},\vv_{t-1})\\
    &=q(\vz_0|\vx_0) q(\vv_0|\vx_0)  \prod_{t=1}^{T} q(\vz_{t}|\vz_{t-1}, \vv_{t-1}) q(\vv_{t}|\vz_{t-1}, \vv_{t-1})\\
    &=q(\vz_0|\vx_0) q(\vv_0|\vx_0)  \prod_{t=1}^{T} \delta(\vz_t - f_\vz(\vz_{t-1},\vv_{t-1}))q(\vv_{t}|\vz_{t-1}, \vv_{t-1})\\
\end{aligned}
$}
\end{equation}
where $q(\vz_{t}|\vz_{t-1}, \vv_{t-1})$ and $q(\vv_{t}|\vz_{t-1}, \vv_{t-1})$ are the successive conditionals for updating $\vz_t$ and $\vv_t$ at each timestep, respectively. Since the joint update of $\vz_t$ and $\vv_t$ is chosen to be autonomous, we omit later $\vx_{t}$ for simplifying above posterior. We derive the lower bound to model evidence (ELBO) as:
\begin{equation}
\resizebox{.90\linewidth}{!}{$
\begin{aligned}
    &\log p(\bar{\vx}) = \E_{q_{\theta}(\bar{\vz},\bar{\vv}|\bar{\vx}))} \left[ \log
\frac{p(\bar{\vx},\bar{\vz},\bar{\vv})}{q(\bar{\vz},\bar{\vv}|\bar{\vx})}
\frac{q(\bar{\vz},\bar{\vv}|\bar{\vx})}{p(\bar{\vz},\bar{\vv}|\bar{\vx})}\right]\\
&\geq \E_{q_{\theta}(\bar{\vz},\bar{\vv}|\bar{\vx}))} \left[ \log\frac{p(\bar{\vx},\bar{\vz},\bar{\vv})}{q(\bar{\vz},\bar{\vv}|\bar{\vx})}\right]\\
&= \E_{q_{\theta}(\bar{\vz},\bar{\vv}|\bar{\vx}))} \left[ \log\frac{p(\bar{\vx},\bar{\vv},\vz_0|\vz_{1:T})}{q(\vz_0,\bar{\vv}|\bar{\vx},\vz_{1:T})}\frac{p(\vz_{1:T})}{q(\vz_{1:T}|\vz_0,\vv_{0:T-1})}\right]\\
&= \E_{q_{\theta}(\bar{\vz},\bar{\vv}|\bar{\vx}))} \left[ \log\frac{p(\bar{\vx},\bar{\vv},\vz_0|\vz_{1:T})}{q(\vz_0,\bar{\vv}|\bar{\vx},\vz_{1:T})}\cancel{\frac{\prod_{t=1}^{T}\delta(\vz_{t} - f_\vz(\vz_{t-1},\vv_{t-1})) }{\prod_{t=1}^{T}\delta(\vz_{t} - f_\vz(\vz_{t-1},\vv_{t-1}))}}\right]\\
&= \E_{q_{\theta}(\bar{\vz},\bar{\vv}|\bar{\vx}))} \left[ \log\frac{p(\bar{\vx}|\bar{\vv},\bar{\vz})p(\bar{\vv})p(\vz_0)}{q(\vz_0)q(\bar{\vv}|\bar{\vx},\bar{\vz})}\right]\\
&= \E_{q_{\theta}(\bar{\vz},\bar{\vv}|\bar{\vx})}\left[ \log p(\bar{\vx}|\bar{\vv},\bar{\vz}) \right]  + \E_{q_{\theta}(\bar{\vz},\bar{\vv}|\bar{\vx})}\left[\log\frac{p(\vz_0)}{q(\vz_0)}\frac{p(\bar{\vv})}{q(\bar{\vv}|\bar{\vx},\bar{\vz})}  \right] 
\end{aligned}
$}
\label{eq:elbo}
\end{equation}
Factorizing the joint distribution over timesteps, we can further re-write the above ELBO as:
\begin{equation}
\resizebox{.90\linewidth}{!}{$
    \begin{gathered}
    \log p(\bar{\vx}) \geq \sum_{t=0}^{T}\E_{q_{\theta}} \big[ \log p(\vx_{t}|\vz_{t},\vv_{t}) \big] \\- \E_{q_{\theta}} \big[\mathrm{D}_{\text{KL}}\left[ q_{\theta}(\vz_0|\vx_0)||p(\vz_0)\right] \big]  - \E_{q_{\theta}} \big[\mathrm{D}_{\text{KL}}\left[ q_{\theta}(\vv_0|\vx_0)||p(\vv_0)\right] \big]\\ -\sum_{t=1}^{T}\E_{q_{\theta}} \big[\mathrm{D}_{\text{KL}}\left[ q_{\theta}(\vv_{t}|\vz_{t-1}, \vv_{t-1}) || p(\vv_{t}) \right] \big]
    \end{gathered}
    $}
\end{equation}
where the first term is the reconstruction objective, the second and third terms define the KL divergence on the initial distribution of latent variables, and the last term regularizes the time evolution of the posterior. 

Recall that we assume that the observed spikes $\bar{\vx}$ are samples from a Poisson process with underlying rates $\bar{\vr}$. At each time step, the firing rates $\vr_t$ are predicted as a function of the latent variables $\vz_t$ and $\vr_t$ from the approximate posterior. The corresponding optimization objectives are defined as:
\begin{equation}
\resizebox{.90\linewidth}{!}{$
\begin{gathered}
    \mathcal{L}^{\text{Poisson}} = - \sum_{t=0}^{T}\E_{q_{\theta}} \big[\log\Big(\text{Poisson}(\vx_t|\vr_t)\Big)  \big]\\
    \mathcal{L}^{KL} =  \E_{q_{\theta}} \big[\mathrm{D}_{\text{KL}}\Big(q_{\theta}(\vz_0|\vx_0)||p(\vz_0)\Big) \big] + \E_{q_{\theta}} \big[\mathrm{D}_{\text{KL}}\Big(q_{\theta}(\vv_0|\vx_0)||p(\vv_0)\Big) \big]\\ +  \sum_{t=1}^{T}\E_{q_{\theta}} \big[\mathrm{D}_{\text{KL}}\Big(q_{\theta}(\vv_{t}|\vz_{t-1}, \vv_{t-1}) || p(\vv_{t})\Big)\big]
    \label{eq:loss}
 \end{gathered}
$}
\end{equation}
where $\mathcal{L}^{\text{Poisson}}$ denotes the Poisson negative log-likelihood, and $\mathcal{L}^{KL}$ represents the KL divergence regularization. In practical implementation, the decoder 
$q$ also incorporates the encoder’s hidden states as part of its input. For brevity, we defer the details and provide a more thorough explanation when introducing the model architecture and referring to Fig.~\ref{fig:arc}.

\subsection{Latent Langevin Posterior Flow}

To derive the time evolution of the posterior, we can decompose the underdamped Langevin equation into two steps:
\begin{equation}
\begin{aligned}
\texttt{Deterministic Step:}\ \ \ \  \frac{\mathop{d\vz}}{\mathop{dt}} &= \vv, \frac{\mathop{d\vv}}{\mathop{dt}} = -\frac{\nabla_\vz U(\vz)}{m}\\
\texttt{Probabilistic Step:}\ \ \ \ \frac{\mathop{d\vv}}{\mathop{dt}} &= - \gamma \vv + \sqrt{2m\gamma k_b\tau}\boldsymbol{\eta}(t)
\label{eq:posterior_flow}
\end{aligned}
\end{equation}
Here the deterministic step amounts to the Hamiltonian flow where the total energy is conserved, and the probabilistic step follows the stochastic Ornstein-Uhlenbeck process. Our goal is to derive the time evolution of joint posterior probability.

\subsubsection{Hamiltonian Flow.} The deterministic step of Eq.~(\ref{eq:posterior_flow}) actually defines the Hamiltonian of the system:
\begin{equation}
    \mathcal{H}(\vv,\vz) = \underbrace{\frac{U(\vz,\vx)}{m}}_{\text{Potential}} + \underbrace{\frac{1}{2}||\vv||^2}_{\text{Kinetic}}
    \label{eq:hamiltonian}
\end{equation}
The total energy is conserved in the time evolution of the coupled variables. Discretizing over time leads to the joint update:
\begin{equation}
\begin{aligned}
    [\vz_{t+\frac{1}{2}},\vv_{t+\frac{1}{2}}] &= f(\vz_t,\vv_t)= [f_\vz, f_\vv]\\ &=[\vz_t+\frac{\partial \mathcal{H}}{\partial \vv_t}\Delta t,\vv_t-\frac{\partial \mathcal{H}}{\partial \vz_t}\Delta t]   
    \label{eq:hamilton_update}
\end{aligned}
\end{equation}
where the subscript $t$ denote the time index, $\Delta t$ represents the step size in physical time, and $[f_{\vz_t}, f_{\vv_t}]$ denotes the above coupled transformation. The joint posterior $q(\vz_{t+\nicefrac{1}{2}},\vv_{t+\nicefrac{1}{2}})$ obeys the normalizing-flow-like density evolution:
\begin{equation}
\resizebox{.85\linewidth}{!}{$
    \log q(\vz_{t+\frac{1}{2}},\vv_{t+\frac{1}{2}}) = \log q(\vz_{t},\vv_{t}) + \log|\det (I + \mathcal{J}_{\mathcal{H}}\Delta t)|^{-1} 
$}
\end{equation}
where $\mathcal{J}_{\mathcal{H}}$ is the Jacobian induced by the Hamiltonian. For infinitesimal steps $\Delta t$, we have:
\begin{equation}
\begin{aligned}
    \det (I + \mathcal{J}_{\mathcal{H}}\mathop{dt}) &= \det \Big[ I + \begin{pmatrix}
        \frac{\partial^2\mathcal{H}}{\partial z_i\partial v_j} & -\frac{\partial^2\mathcal{H}}{\partial z_i\partial z_j} \\
        \frac{\partial^2\mathcal{H}}{\partial v_i\partial z_j} & -\frac{\partial^2\mathcal{H}}{\partial v_i\partial z_j}
    \end{pmatrix}\Delta t\Big]\\
    &\approx 1 + \Tr \begin{pmatrix}
        \frac{\partial^2\mathcal{H}}{\partial z_i\partial v_j} & -\frac{\partial^2\mathcal{H}}{\partial z_i\partial z_j} \\
        \frac{\partial^2\mathcal{H}}{\partial v_i\partial z_j} & -\frac{\partial^2\mathcal{H}}{\partial v_i\partial z_j}
    \end{pmatrix}\Delta t\approx 1 
\end{aligned}
\end{equation}
The deterministic conditional can thus be written as:
\begin{equation}
\resizebox{.85\linewidth}{!}{$
    \begin{aligned}
        q(\vz_{t+\frac{1}{2}},\vv_{t+\frac{1}{2}}|\vz_{t},\vv_{t})
        \approx \delta\Big(\vz_{t+\frac{1}{2}} - f_\vz(\vz_t,\vv_t),\vv_{t+\frac{1}{2}} - f_\vv(\vz_t,\vv_t)\Big)
    \end{aligned}
$}
\end{equation}
The posterior conserves probability mass over time. Let $q(\vz_{t+1}|\vz_{t+\nicefrac{1}{2}})$ denote a trivial Dirac $\delta$-function. Marginalizing $\vz_{t+1/2}$ out gives the conditional of $\vz_{t+1}$:
\begin{equation}
\begin{aligned}
    &q(\vz_{t+1}|\vz_t,\vv_t) = \int q(\vz_{t+1}|\vz_{t+\frac{1}{2}})q(\vz_{t+\frac{1}{2}}|\vz_{t},\vv_{t})\mathop{d\vz_{t+\frac{1}{2}}}\\
    &=\int\delta\Big(\vz_{t+1} - \vz_{t+\frac{1}{2}}\Big)\delta\Big(\vz_{t+\frac{1}{2}} - f_\vz(\vz_t,\vv_t)\Big)\mathop{d\vz_{t+\frac{1}{2}}}\\
    &=\delta\Big(\vz_{t+1} - f_\vz(\vz_t,\vv_t)\Big)
\end{aligned}
\end{equation}

\subsubsection{Ornstein-Uhlenbeck Process.} The probabilistic step of Eq.~(\ref{eq:posterior_flow}) is given by the Ornstein–Uhlenbeck process which describes a noisy relaxation process, whereby a particle is disturbed with noise $\eta(t)$ and simultaneously relaxed to its mean position with friction coefficient $\gamma$: 
\begin{equation}
    \frac{\mathop{d\vv}}{\mathop{dt}} =  -\gamma \vv  + \sqrt{2m\gamma k_B \tau}\ \eta(t).
    \label{eq:brownian_dynamics}
\end{equation}
Discretizing over timesteps, the Gaussian noise yields a Gaussian transition update as:
\begin{equation}
\resizebox{.90\linewidth}{!}{$
\begin{aligned}
    &q(\vv_{t+1}|\vz_{t}, \vv_{t}) = \int q(\vv_{t+1}| \vv_{t+\frac{1}{2}})q(\vv_{t+\frac{1}{2}}|\vz_t,\vv_t)\mathop{d\vv_{t+\frac{1}{2}}}\\
    &=\int\mathcal{N}\Big((1-\gamma)\vv_{t+\frac{1}{2}}, 2m\gamma k_B \tau I\Big)\delta\Big(\vv_{t+\frac{1}{2}} - f_\vv(\vz_t,\vv_t)\Big)\mathop{d\vv_{t+\frac{1}{2}}}\\
    &=\mathcal{N}\Big((1-\gamma)f_\vv(\vz_t,\vv_t), 2m\gamma k_B \tau I\Big)
\end{aligned}
$}
\label{eq:gaussian_tran}
\end{equation}
The re-parameterization trick~\citep{vae} is used to allow for differentiation through the Gaussian kernel. We alternate these two steps to compute the conditional update of the joint posterior $q(\vz_{t+1},\vv_{t+1}|\vz_{t},\vv_{t})$. 

\begin{figure}[htbp]
    \centering
    \includegraphics[width=0.99\linewidth]{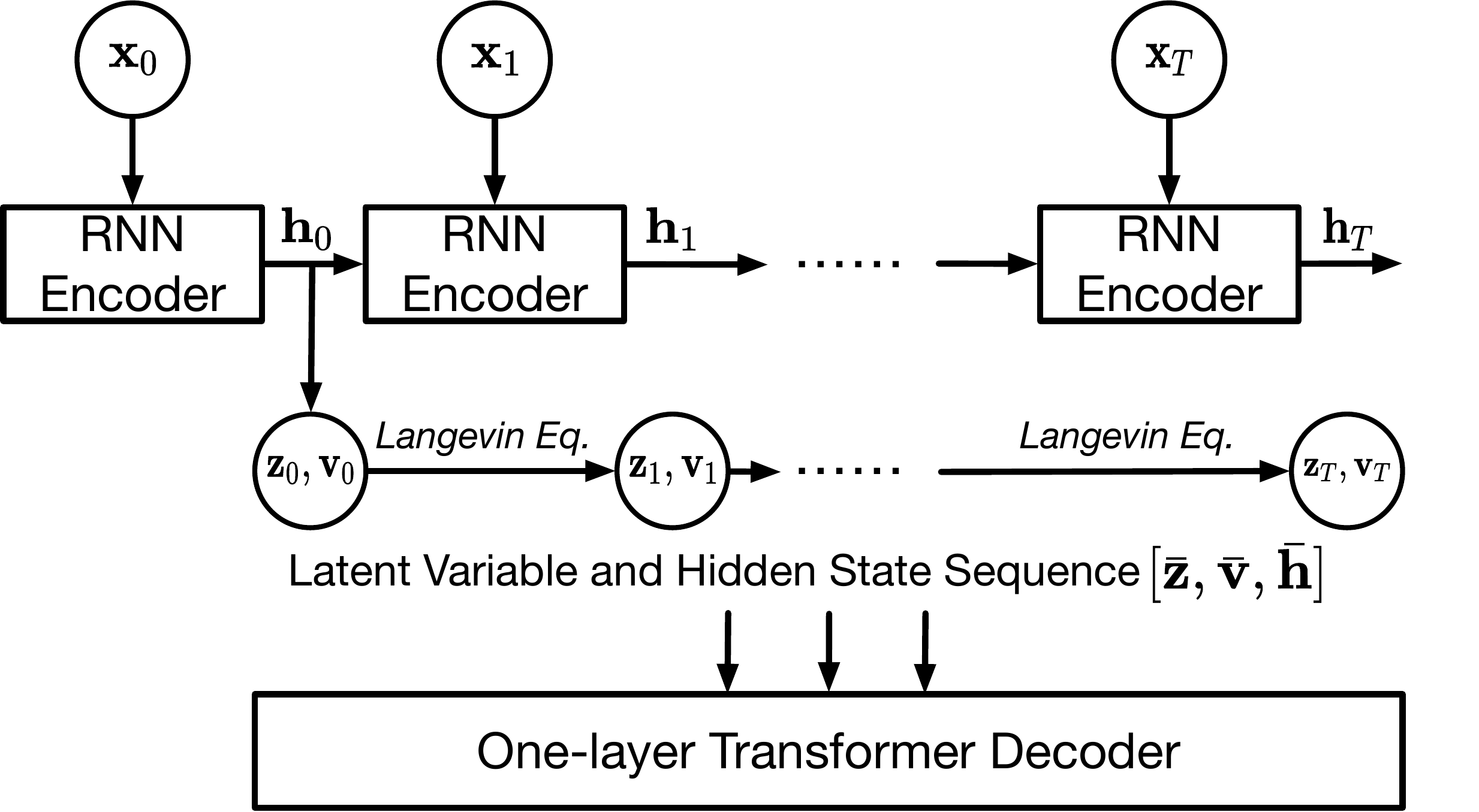}
    \caption{Workflow of our method: the RNN encoder takes the spike data as input at every timestep and updates the hidden states $\vh_t$, and the latent variables $\vz_t,\vv_t$ evolve in time according to the Langevin equation. Finally, the Transformer decoder predicts the firing rates from the entire sequence.}
    \label{fig:arc}
\end{figure}

\begin{algorithm}[htbp]
\caption{Training algorithm of our Langevin flow.}
\begin{algorithmic}[1]
\label{alg:training}
\REQUIRE Recurrent encoder $\texttt{GRU}$, Transformer-based sequence decoder $\texttt{Transformer}$, linear mapping for latent variables $n$, input spike sequence $\bar{\vx}$, and posterior $q_\theta$.\\
\REPEAT
\STATE Initial hidden states: $\vh_0 = \texttt{GRU}(\vx_0)$ \\
\STATE Initial latent variables: $[\vz_0, \vv_0] = n(\vh_0)$
\STATE Time step counter: $i=0$ \\
\WHILE{$i\leq T-1$}
\STATE Update position (deterministic step): $\vz_{i+1} = \vz_{i} + \vv_{i}$
\STATE Update velocity (deterministic step): 
\\ $\vv_{i+\frac{1}{2}} = \vv_{i} - \nicefrac{\nabla_\vz U(\vz_{i})}{m}$
\STATE Update velocity (probabilistic step): \\ $ \vv_{i+1} = (1-\gamma) \vv_{i+\frac{1}{2}}  + \sqrt{2m\gamma k_B \tau}\  \boldsymbol{\eta}(i)$
\STATE Update hidden states: $\vh_{i+1} = \texttt{GRU}(\vx_{i+1},\vh_i)$
\STATE Concatenate variable sequences: \\ $\bar{\vz}=[\vz_{0:i},\vz_{i+1}],\  \bar{\vv}=[\vv_{0:i},\vv_{i+1}],\  \bar{\vh}=[\vh_{0:i},\vh_{i+1}]$
\STATE Update time step counter: $i=i+1$
\ENDWHILE
\STATE Predict firing rates: $\bar{\vr} = \texttt{Transformer}(\bar{\vz},\bar{\vv},\bar{\vh})$
\STATE Optimize the $ \mathcal{L}^{\text{Poisson}}$ and $\mathcal{L}^{KL}$. 
\UNTIL converged
\end{algorithmic}
\end{algorithm}

\begin{figure*}[t]
    \centering
    \includegraphics[width=0.9\linewidth]{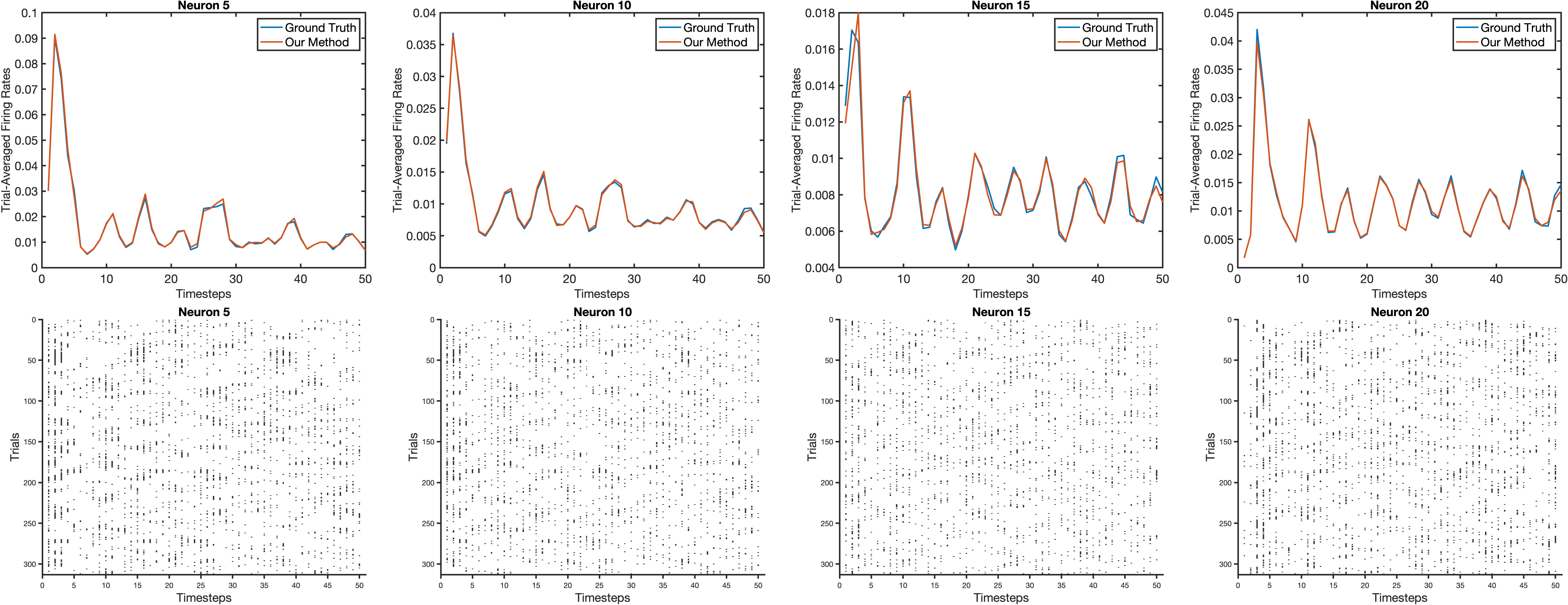}
    \vspace{-3mm}
    \caption{Trial-average firing rates (top) and the corresponding spike trains (bottom) of some neurons of Lorenz system.}
    \label{fig:Lorenz_rates_spikes}
\end{figure*}

\begin{table*}[tbp]
    \centering
    \vspace{-5mm}
    \caption{Results on \texttt{MC\_Maze} and \texttt{MC\_RTT} with the sampling frequency of $20$ ms. }
    \resizebox{0.99\linewidth}{!}{
    \begin{tabular}{c|cccc|ccc}
    \toprule
        \multirow{2}*{Methods} & \multicolumn{4}{c|}{MC-Maze} &\multicolumn{3}{c}{MC-RTT} \\
        & co-bps ($\uparrow$) & vel R2 ($\uparrow$) & psth R2 ($\uparrow$) & fp-bps ($\uparrow$)& co-bps ($\uparrow$) & vel R2 ($\uparrow$) & fp-bps ($\uparrow$)\\
        \midrule
        Smoothing~\citep{yu2008gaussian} & 0.2076	&0.6111	&-0.0005 & -- & 0.1454&0.3875&--\\
        GPFA~\citep{yu2008gaussian} &0.2463&0.6613&0.5574 & -- & 0.1769&0.5263&--\\
        SLDS~\citep{linderman2017bayesian}& 0.2117&0.7944&0.4709&-0.1513 & 0.1662&0.5365&-0.0509\\
        NDT~\citep{ye2021representation} & \textbf{\textcolor{black!60}{0.3597}}&0.8897&0.6172&0.2442& 0.1643&0.6100&0.1200\\
        AutoLFADS~\citep{pandarinath2018inferring} &0.3554&0.8906&0.6002&\textbf{\textcolor{black!60}{0.2454}} &0.1976&0.6105&\textbf{\textcolor{black!60}{0.1241}}\\
        MINT~\citep{perkins2023simple} &0.3295&\textbf{0.9005}&\textbf{0.7474}&0.2076 & \textbf{\textcolor{black!60}{0.2008}}&\textbf{\textcolor{black!60}{0.6547}}&0.1099\\
        \rowcolor{gray!40} LangevinFlow & \textbf{0.3641}&\textbf{\textcolor{black!60}{0.8940}}&\textbf{\textcolor{black!60}{0.6801}}&\textbf{0.2573}& \textbf{0.2010}&\textbf{0.6652}&\textbf{0.1389}\\
    \bottomrule
    \end{tabular}
    }
    \label{tab:nlb_20ms_maze_rtt}
\end{table*}

\subsection{Architecture and Training Algorithm}
\label{sec:architecture}

Fig.~\ref{fig:arc} displays our model architecture. A recurrent encoder $\texttt{GRU}$~\citep{chung2014empirical} is used to encode the input sequence to a set of hidden states $\mathbf{h}_t = \texttt{GRU}(\mathbf{x}_{t-1}, \mathbf{h}_{t-1})$. The initial conditions for the latent variables $\mathbf{z}_0 \ \& \ \mathbf{v}_0$ are inferred from $\mathbf{h}_0$, and then evolve forward in time according to both the deterministic and stochastic steps. The RNN encoder is included to model the short-range dependencies of neural data. After encoding input spikes and performing latent Langevin flow, all hidden states and latent variables are combined through a single Transformer~\citep{waswani2017attention} layer to predict the firing rates of the sequence: $\mathbf{\bar{r}} = \texttt{Transformer}(\mathbf{\bar{z}}, \mathbf{\bar{v}}, \mathbf{\bar{h}})$. We use a Transformer for decoding because it can capture long-range interactions over time and allows for a more globally informed prediction of firing rates. The parameters of the GRU, Transformer, linear readout, and potential are then optimized to maximize the ELBO in Eq.~(\ref{eq:elbo}). We summarize the training algorithm in Alg.~\ref{alg:training}.

\begin{table*}[thbp]
    \centering
    \caption{Results on \texttt{Area2\_Bump} and \texttt{DMFC\_RSG} with the sampling frequency of $20$ ms. }
    \resizebox{0.99\linewidth}{!}{
    \begin{tabular}{c|cccc|cccc}
    \toprule
        \multirow{2}*{Methods} &\multicolumn{4}{c|}{Area2-Bump}&\multicolumn{4}{c}{DMFC-RSG} \\
        &co-bps ($\uparrow$) & vel R2 ($\uparrow$) & psth R2 ($\uparrow$) & fp-bps ($\uparrow$)& co-bps ($\uparrow$) & tp corr ($\downarrow$) & psth R2 ($\uparrow$) & fp-bps ($\uparrow$)\\
         \midrule
        Smoothing~\citep{yu2008gaussian} &0.1529&0.5319&-0.1840&-- & 0.1183&-0.5115&0.2830 &--\\
        GPFA~\citep{yu2008gaussian}  & 0.1791&0.6094&0.5998 &-- & 0.1378&-0.5506&0.3180 &--\\
        SLDS~\citep{linderman2017bayesian} & 0.1816&0.6967&0.5200&0.0132 & 0.1575&-0.5997&0.5470&0.0374\\
        NDT~\citep{ye2021representation} & 0.2624&0.8623&0.6078&0.1459 & 0.1757&-0.6928&0.5477&0.1649\\
        AutoLFADS~\citep{pandarinath2018inferring} & 0.2542&0.8565&0.6552&0.1423&\textbf{\textcolor{black!60}{0.1871}}&\textbf{-0.7819}&0.5903&\textbf{\textcolor{black!60}{0.1791}}\\
        MINT~\citep{perkins2023simple} &\textbf{\textcolor{black!60}{0.2718}}&\textbf{\textcolor{black!60}{0.8803}}&\textbf{0.9049}&\textbf{\textcolor{black!60}{0.1489}} &0.1824&\textbf{\textcolor{black!60}{-0.6995}}&\textbf{0.7014}&0.1647\\
        \rowcolor{gray!40} LangevinFlow & \textbf{0.2881}&\textbf{0.8810}&\textbf{\textcolor{black!60}{0.7641}}&\textbf{0.1647} & \textbf{0.1904}&-0.5981&\textbf{\textcolor{black!60}{0.6079}}&\textbf{0.1945}\\
    \bottomrule
    \end{tabular}
    }
    \label{tab:nlb_20ms_bump_rsg}
\end{table*}

 \begin{figure*}[tbp]
    \centering
    \includegraphics[width=0.88\linewidth]{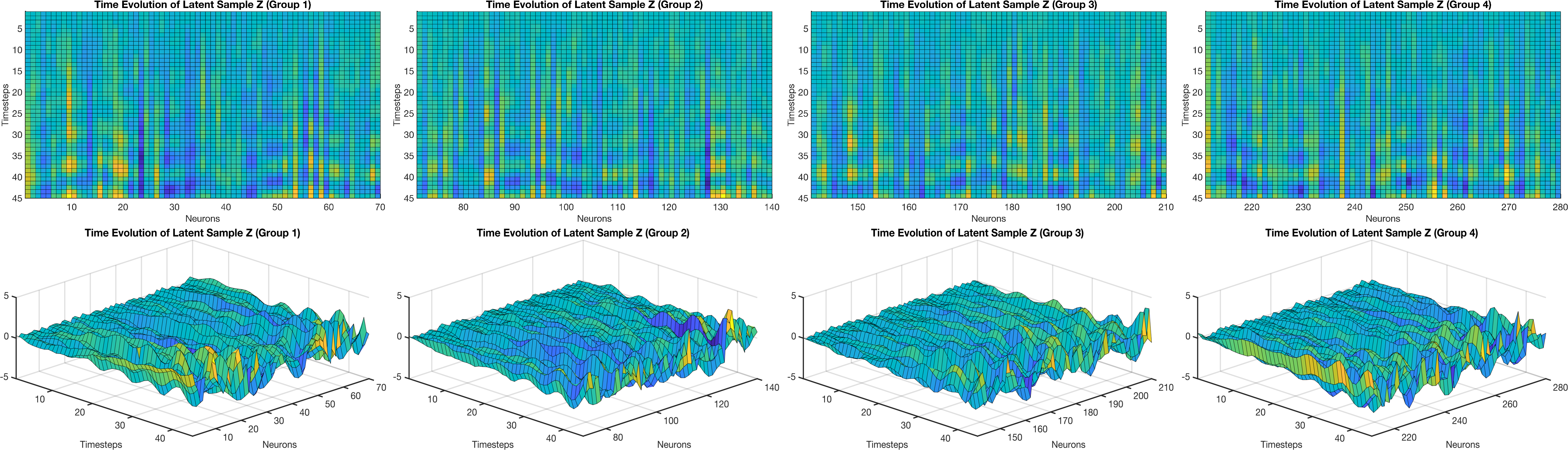}
    \caption{Spatiotemporal waves induced by our LangevinFlow in different views on \texttt{MC\_Maze}. Here each group denotes an independent set of convolution channels.}
    \label{fig:waves}
\end{figure*}

\section{Experiments}

This section presents the experimental setup and the results. We start with the setup of the experiments, discuss the results on the toy dataset of Lorenz attractor, and finally present the extensive evaluation of the Neural Latents Benchmark. 

\subsection{Setup}


\subsubsection{Baselines.} On the synthetic Lorenz attractor dataset, we mainly compare with AutoLFADS~\citep{autolfads} and NDT~\citep{ye2021representation}, which are dedicated RNN and Transformer architectures designed for neural population modeling. On NLB, we further compare with a wide range of competitive baselines on the public leaderboard \footnote{\url{https://eval.ai/web/challenges/challenge-page/1256/leaderboard/}}. 

\subsubsection{Implementation Details.} The Transformer decoder consists of only $1$ self-attention layer with $4$ attention heads. Another linear layer is used for reading out firing rates. For the Langevin equation, the mass $m$ is set to an identity matrix, and both Boltzman constant $k_B$ and temperature $\tau$ are set $1$. The damping ratio $\gamma$ is tuned for specific datasets but stays in the range of $[0.55,0.8]$. For the potential, the latent code is first divided into $4$ groups (\emph{i.e.,} independent convolution channels ), and we use a one-dimensional convolution layer of kernel size $7$ with padding $3$ and stride $1$ for each group. We adopt a hyper-parameter $\lambda$ to tune the strength of the KL penalty and add a scheduler to gradually increase the value so that the optimization does not quickly set the KL divergence to $0$. As the observed spikes are assumed to be from a low-dimensional subspace, we use coordinated dropout~\citep{keshtkaran2019enabling} to randomly drop input samples during the training, which enforces the model to learn the underlying latent structure shared across neurons. 

\subsection{Synthetic Lorenz Attractor} 

The Lorenz attractor is a 3D dynamical system where the dynamics are governed by three coupled non-linear equations:
\begin{equation}
\begin{aligned}
     \dot{y}_1 &= \sigma (y_2 - y_3),\\
     \dot{y}_2 &= y_1 (\rho-y_3) y_2,\\
     \dot{y}_3 &= y_1y_2 -\beta y_3
\end{aligned}
\end{equation}
where $\sigma,\rho,\beta$ are hyper-parameters. In line with~\citet{ye2021representation}, we first simulate the 3D Lorenz attractor and then project the 3D states into a higher dimensionality using a random linear transform to form firing rates for a population of synthetic neurons. The spikes of each trial are sampled from the Poisson distribution with these firing rates. The evaluating methods are expected to infer the true firing rates of the Lorenz system from the synthetic spiking activity alone.

\begin{table}[htbp]
    \centering
    \caption{$R_2$ of the firing rates on Lorenz Attractor.}
    \begin{tabular}{c|ccc}
    \toprule
         & AutoLFADS & NDT & \cellcolor{gray!40}LangevinFlow\\
    \midrule
       $R_2 (\uparrow)$ & 0.921$\pm$0.005 &0.934$\pm$0.004 & \cellcolor{gray!40}\textbf{0.944$\pm$0.003}\\
    \bottomrule
    \end{tabular}
    \label{tab:Lorenz}
\end{table}


Table~\ref{tab:Lorenz} presents the results of $R_2$ correlation between the predicted firing rates and the ground truth. Our LangevinFlow outperforms the baselines and achieves a higher correlation score in predicting the firing rates, indicating that our model more accurately captures the underlying dynamical structure of the synthetic neural data. Fig.~\ref{fig:Lorenz_rates_spikes} compares the predicted trial-averaged firing rates and of several randomly selected neurons alongside their ground truth counterparts, as well as the corresponding spike trains. Our method closely recovers the general shape and amplitude of the firing rate curves, and also accurately reflects the temporal structure of spike trains.

\begin{table*}[tbp]
    \centering
    \caption{Results on \texttt{MC\_Maze} and \texttt{MC\_RTT} with the sampling frequency of $5$ ms. }
    \resizebox{0.99\linewidth}{!}{
    \begin{tabular}{c|cccc|ccc}
    \toprule
        \multirow{2}*{Methods} & \multicolumn{4}{c|}{MC-Maze} &\multicolumn{3}{c}{MC-RTT} \\
        & co-bps ($\uparrow$) & vel R2 ($\uparrow$) & psth R2 ($\uparrow$) & fp-bps ($\uparrow$)& co-bps ($\uparrow$) & vel R2 ($\uparrow$) & fp-bps ($\uparrow$)\\
        \midrule
        Smoothing~\citep{yu2008gaussian} & 0.2109&0.6238&0.1853 & -- & 0.1468&0.4142&--\\
        GPFA~\citep{yu2008gaussian} &0.1872	&0.6399	&0.5150 & -- & 0.1548	&0.5339&--\\
        SLDS~\citep{linderman2017bayesian}& 0.2249&0.7947&0.5330&-0.1513 & 	0.1649&0.5206&0.0620\\
        NDT~\citep{ye2021representation} & 0.3229&0.8862&0.5308&0.2206& 0.1749	&0.5656&0.0970\\
        AutoLFADS~\citep{pandarinath2018inferring} &\textbf{\textcolor{black!60}{0.3364}}&\textbf{\textcolor{black!60}{0.9097}}&\textbf{\textcolor{black!60}{0.6360}}&\textbf{\textcolor{black!60}{0.2349}}&0.1868&\textbf{\textcolor{black!60}{0.6167}}&\textbf{\textcolor{black!60}{0.1213}}\\
        MINT~\citep{perkins2023simple} &0.3304&\textbf{0.9121}&\textbf{0.7496}&0.2076& \textbf{0.2014}	&\textbf{0.6559}&0.1099\\
        \rowcolor{gray!40} LangevinFlow &\textbf{0.3624}&0.7867&0.5515&\textbf{0.2556}&\textbf{\textcolor{black!60}{0.1900}}&0.4748&\textbf{0.1300}\\
    \bottomrule
    \end{tabular}
    }
    \label{tab:nlb_5ms_maze_rtt}
\end{table*}
\begin{table*}[tbp]
    \centering
    \caption{Results on \texttt{Area2\_Bump} and \texttt{DMFC\_RSG} with the sampling frequency of $5$ ms. }
    \resizebox{0.99\linewidth}{!}{
    \begin{tabular}{c|cccc|cccc}
    \toprule
        \multirow{2}*{Methods} &\multicolumn{4}{c|}{Area2-Bump}&\multicolumn{4}{c}{DMFC-RSG} \\
        &co-bps ($\uparrow$) & vel R2 ($\uparrow$) & psth R2 ($\uparrow$) & fp-bps ($\uparrow$)& co-bps ($\uparrow$) & tp corr ($\downarrow$) & psth R2 ($\uparrow$) & fp-bps ($\uparrow$)\\
         \midrule
        Smoothing~\citep{yu2008gaussian}  &0.1544&0.5736&0.2084&--&0.1202&	-0.5139&0.2993&--\\
        GPFA~\citep{yu2008gaussian}   &0.1680&0.5975&0.5289&--&0.1176&-0.3763	&0.2142&--\\
        SLDS~\citep{linderman2017bayesian}  &0.1960&0.7385&0.5740&0.0242&0.1243&-0.5412&0.3372&-0.0418\\
        NDT~\citep{ye2021representation}  &0.2623&\textbf{\textcolor{black!60}{0.8672}}&0.6619&0.1184&0.1720	&-0.5624&0.4377&0.1404\\
        AutoLFADS~\citep{pandarinath2018inferring}  &0.2569&0.8492&0.6318&\textbf{\textcolor{black!60}{0.1505}}&\textbf{\textcolor{black!60}{0.1829}}&\textbf{-0.8248}&\textbf{\textcolor{black!60}{0.6359}}&\textbf{0.1844}\\
        MINT~\citep{perkins2023simple}  &\textbf{\textcolor{black!60}{0.2735}}&\textbf{0.8877}&\textbf{0.9135}&0.1483&0.1821	&\textbf{\textcolor{black!60}{-0.6929}}&\textbf{0.7013}&0.1650\\
        \rowcolor{gray!40} LangevinFlow &\textbf{0.2772}&0.8580&\textbf{\textcolor{black!60}{0.7567}}&\textbf{0.1526}&\textbf{0.1841}&-0.5466&0.6092&\textbf{\textcolor{black!60}{0.1689}}\\
    \bottomrule
    \end{tabular}
    }
    \label{tab:nlb_5ms_bump_rsg}
\end{table*}

\begin{figure}[htbp]
    \centering
    \includegraphics[width=0.99\linewidth]{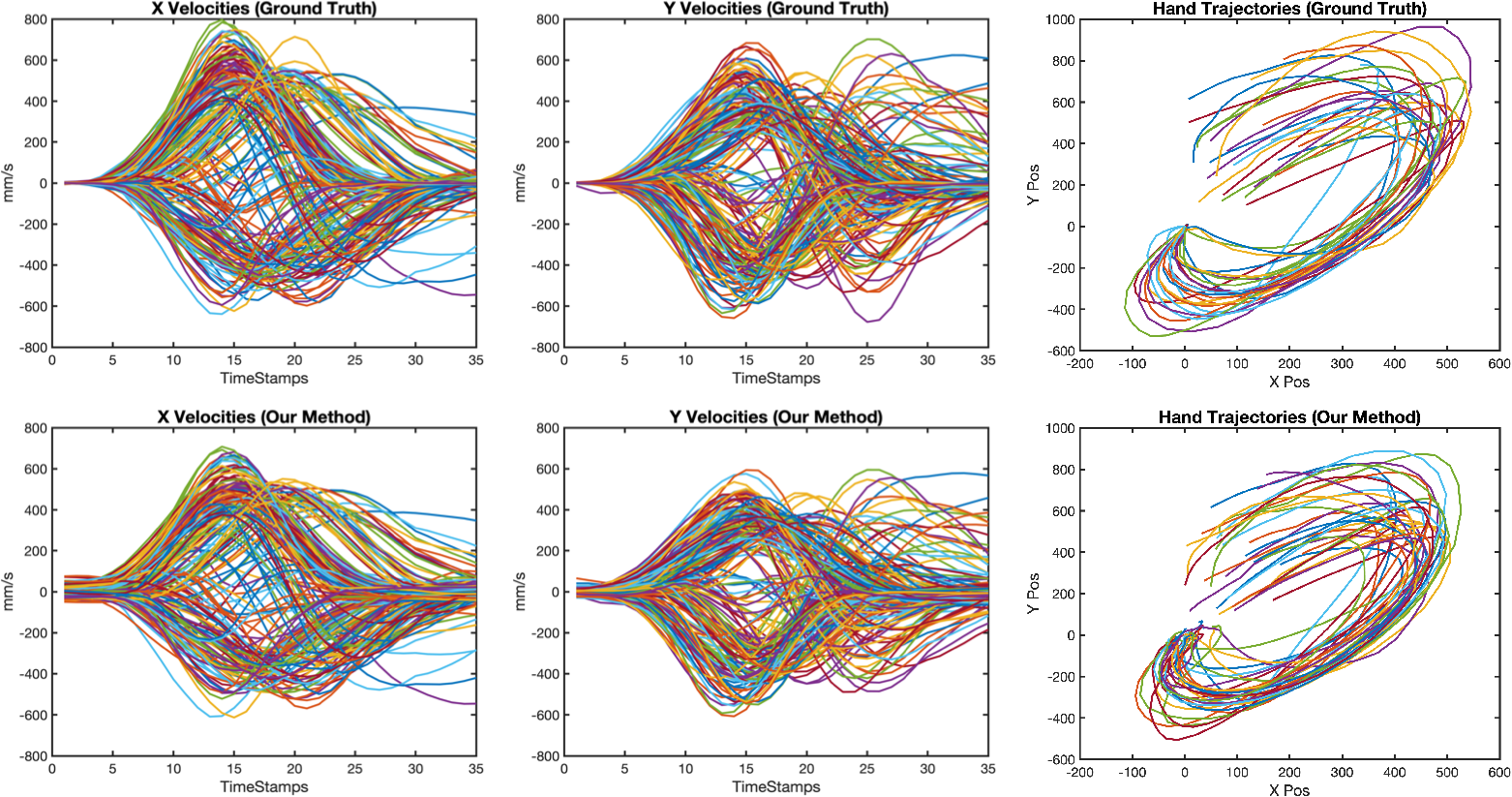}
    \caption{Kinematics (hand velocities and trajectories) of the ground truth and predicted by our method on \texttt{Area2\_Bump}. }
    \label{fig:kinematics}
\end{figure}

\subsection{Neural Latent Benchmark}

The NLB~\citep{pei2021neural} is a benchmark designed for evaluating unsupervised approaches that model neural population activities. This benchmark provides four curated neuro-physiological datasets from monkeys that span motor, sensory, and cognitive brain regions, with behaviors that vary from pre-planned, stereotyped movements to those in which sensory input must be dynamically integrated. The primary metric \texttt{co-smoothing}~\citep{macke2011empirical} evaluates the normalized log-likelihood of held-out neuronal activity prediction, while the secondary metrics can include behavior decoding accuracy, match to PSTH, or log-likelihood of forward predictions. Two sampling frequencies ($5$ms and $20$ms) are pre-defined to obtain datasets with different sequence lengths.

\subsubsection{20 ms Results.} In Table \ref{tab:nlb_20ms_maze_rtt} and~\ref{tab:nlb_20ms_bump_rsg}, we see that our LangevinFlow achieves state of the art on the likelihood of held-out neurons (co-smoothing bits per spike), as well as forward prediction bits per spike. The model also compares very favorably in terms of the behavioral metrics such as hand-velocity regression. The model is not state of the art on PSTH $R^2$, which may be expected given the known trade-off between the co-bps metric and the performance on trial-averaged PSTH correlation metric. The overall performance across multiple metrics underscores its robustness in capturing neural dynamics.


\subsubsection{5 ms Results.} Table~\ref{tab:nlb_5ms_maze_rtt} and~\ref{tab:nlb_5ms_bump_rsg} report the evaluation results on NLB with the sampling frequency of $5$ ms. The results at this higher temporal resolution are very coherent with those of $20$ ms. Our LangevinFlow maintains impressive performance, achieving strong likelihood scores on held-out neurons and forward predictions on most datasets. This consistency across different sampling frequencies confirms the model’s ability to adapt to varying temporal granularities, which is critical for capturing the fine-scale dynamics present in neural data.

\subsubsection{Spatiotemporal Wave Dynamics.} Fig.~\ref{fig:waves} displays the smooth spatiotemporal waves induced by our coupled oscillator potential. We can see that the latent variables in different convolution groups exhibit clear but distinct wave patterns, reminiscent of traveling waves observed in cortical activity~\citep{ermentrout2001traveling, muller2014stimulus}. Such wave dynamics are thought to play several key computational roles in neuroscience. For example, traveling waves have been proposed to facilitate the integration of information over distributed neural populations, serving as a mechanism for coordinating activity across different brain regions~\citep{buzsaki2006rhythms, miller2009power,muller2018cortical}. They can help synchronize the timing of neural firing, thereby enhancing signal propagation and ensuring that information is efficiently routed and integrated. Moreover, wave dynamics may support processes such as working memory, decision-making, preditive decoding, and sensorimotor integration~\citep{sato2012traveling,engel2013intrinsic,besserve2015shifts,alamia2019alpha,friston2019waves}. In our model, the emergence of these wave patterns not only reflects the inherent oscillatory dynamics of the neural data, but also suggests that our coupled oscillator potential may be capturing similar computational principles, contributing to the robust performance of our approach.


\subsubsection{Kinematics Visualization.} Fig.~\ref{fig:kinematics} illustrates the time evolution of key kinematic variables on the \texttt{Area2\_Bump} dataset, including hand X and Y velocities, as well as the overall hand trajectories. The high behavior decoding accuracy of our model is evident here: linear regression models fitted on predicted firing rates yield kinematic outputs that closely match the ground truth. These results validate the accuracy of the neural activity reconstruction, demonstrating the practical utility of our approach in decoding behaviorally relevant signals. 



\subsubsection{Ablation Studies.} Finally, we designed a number of baselines and performed ablations to understand the role each component of our LangevinFlow plays on overall performance. Specifically, we considered the following model variants:
\begin{itemize}
    \item \textit{Baseline 1}: a linear decoder in place of the Transformer.
    \item \textit{Baseline 2}: a linear encoder with no hidden states.
    \item \textit{Baseline 3}: a model without Langevin dynamics relying solely on hidden state dynamics.
    \item  \textit{Baseline 4}: a variant in which the oscillator potential also couples latent variables to input spikes. Explicitly:  $U(\vz,\vx) = \vz^T \frac{\mW_\vz}{||\mW_\vz||_2} \vz  + \vz^T\mW_\vx\vx$.
    \item \textit{Baseline 5}: a version using first-order dynamics instead of Langevin dynamics. Explicitly: $\vz_{t+1} = \vz_t - \nabla_{\vz_t} U(\vz_t)$.
\end{itemize}    
Table~\ref{tab:ablation} shows the results on \texttt{MC\_Maze} and \texttt{Area2\_Bump}. In \textit{Baseline 1}, we replaced the Transformer decoder with a linear decoder. Compared to LangevinFlow, this variant has slightly lower co-smoothing (co-bps) and forward prediction (fp-bps) scores, which indicate the importance of the Transformer in capturing global interactions across the entire latent sequence and in refining firing rate predictions. The global attention mechanism appears to integrate information more effectively than a simpler linear mapping.

\textit{Baseline 2} removes the hidden states from the encoder by replacing the recurrent network with a linear encoder. This modification leads to a noticeable drop in performance, particularly in the co-bps and velocity R2 scores. This suggests that the local temporal dependencies captured by recurrent hidden states are essential for modeling the short-range dynamics present in the neural spike activity.

\textit{Baseline 3} completely omits Langevin dynamics and relies solely on hidden state dynamics. This modification results in a marked performance drop, especially evident in the significant drop in PSTH R2 on \texttt{MC\_Maze}. This decline emphasizes the crucial role of incorporating Langevin dynamics with a learned potential, which represents intrinsic autonomous processes and facilitates the emergence of oscillations. The Langevin dynamics are thus expected to help the model capture the underlying dynamical system more faithfully.

In \textit{Baseline 4}, we augment the oscillator potential by incorporating the input spiking signal. This modification does not provide a substantial benefit to the performance and in some cases slightly underperforms the original model. This result suggests that the learned potential function in its original formulation is already capturing the necessary dependencies.

Finally, \textit{Baseline 5} substitutes the second-order Langevin dynamics with a simpler first-order update rule. The observed performance drop in several metrics confirms that the second-order Langevin dynamics -- featuring terms for inertia and damping -- is more effective in modeling the neural dynamics. The richer dynamics afforded by the second-order formulation appear to better capture both the smooth evolution and the inherent variability of the underlying latent factors.
\begin{table}[htbp]
    \centering
    \caption{Results of ablation studies with the sampling frequency of $20$ ms on \texttt{MC\_Maze} (top) and \texttt{Area2\_Bump} (bottom).}
    \resizebox{0.99\linewidth}{!}{
    \begin{tabular}{c|cccc}
    \toprule
        & co-bps ($\uparrow$) & vel R2 ($\uparrow$) & psth R2 ($\uparrow$) & fp-bps ($\uparrow$)\\
        \midrule
        LangevinFlow & \textbf{0.3641}&0.8940&\textbf{0.6801}&\textbf{0.2573}\\
        \midrule
        Baseline 1 &0.3572&0.8893&0.6683&0.2419\\
        Baseline 2 &0.3328&0.8579&0.6812&0.2549\\
        Baseline 3 &0.3441&0.8109&0.4684&0.2506\\
        Baseline 4 &0.3612&\textbf{0.9005}&0.6743&0.2469\\
        Baseline 5 &0.3586&0.8932&0.6881&0.2351\\
        \bottomrule
        \end{tabular}
        }
    \resizebox{0.99\linewidth}{!}{
    \begin{tabular}{c|cccc}
        & co-bps ($\uparrow$) & vel R2 ($\uparrow$) & psth R2 ($\uparrow$) & fp-bps ($\uparrow$)\\
        \midrule
        LangevinFlow &\textbf{0.2881}&\textbf{0.8810}&0.7641&\textbf{0.1647}\\
        \midrule
        Baseline 1 &0.2795&0.8725&0.7386&0.1549\\
        Baseline 2 &0.2679&0.8641&0.7013&0.1488\\
        Baseline 3 &0.2838&0.8552&0.7165&0.1498\\
        Baseline 4 &0.2800&0.8739&\textbf{0.7803}&0.1631\\
        Baseline 5 &0.2843&0.8614&0.6994&0.1596\\
        \bottomrule
        \end{tabular}
        }
        \label{tab:ablation}
    \end{table}

\section{Conclusions} 

This paper presents LangevinFlow, a sequential variational autoencoder whose latent dynamics are governed by underdamped Langevin equations. By embedding physically grounded stochastic processes and coupled oscillatory behavior into the latent space, our framework offers a powerful avenue for modeling complex neural population activity. We anticipate that these ideas will inspire further exploration of physics-informed inductive biases in neural latent variable modeling, paving the way for even richer and more interpretable dynamical systems approaches.

\noindent\textbf{Limitations and Future Work.} While our framework was shown to yield very promising results, our proposed Langevin dynamics with the present potential function operate in a largely autonomous manner. This formulation seemed to work better than an input-dependent potential in ablation studies; however, adding more input dependence to this potential should intuitively help Langevin dynamics better account for external influences. In future work, exploring more complex input-dependent potential functions could likely yield significant benefits and are a promising new avenue for research uniquely enabled by our LangevinFlow framework.

\section*{Acknowledgments}

This research was supported in part by gifts from Cisco and OpenAI.

\bibliographystyle{ccn_style}

\bibliography{ccn_style}

\end{document}